\documentclass[10pt,twocolumn,a4paper]{article}

\usepackage{times}
\usepackage{epsfig}
\usepackage{graphicx}
\usepackage{amsmath}
\usepackage{amssymb}
\usepackage{multirow}
\usepackage{url}
\usepackage{authblk}
\usepackage{tabularx}
\usepackage{booktabs}

\usepackage[pagebackref=true,breaklinks=true,letterpaper=true,colorlinks,bookmarks=false]{hyperref}
\usepackage{xspace}

\newcommand*{\eg}{e.g.\@\xspace}
\newcommand*{\ie}{i.e.\@\xspace}
\newcommand*{\etal}{et~al.\@\xspace}

\usepackage{cleveref}
\newcommand{\name}{WSGN}
\newcommand{\nws}{Naive Weak Supervision (\cref{eq.naive})}
\newcommand{\nwss}{Naive W. Sup. (\cref{eq.naive})}

\begin{document}
	
	\title{Weakly Supervised Gaussian Networks for Action Detection}
	

\author[]{Basura Fernando}
\author[]{Cheston Tan Yin Chet}
\author[]{Hakan Bilen}

\affil[]{Human Centric AI Programme, A*STAR Artificial Intelligence Initiative (A*AI), A*STAR, Singapore}
\affil[]{I2R, A*STAR, Singapore}
\affil[]{VICO, University of Edinburgh, United Kingdom}

	\date{}
	
	\maketitle

	\begin{abstract}	
		Detecting temporal extents of human actions in videos is a challenging computer vision problem that requires detailed manual supervision including frame-level labels. 
		This expensive annotation process limits deploying action detectors to a limited number of categories. 
		We propose a novel method, called WSGN, that learns to detect actions from \emph{weak supervision}, using only video-level labels. 
		WSGN learns to exploit both video-specific and dataset-wide statistics to predict relevance of each frame to an action category. 
		This strategy leads to significant gains in action detection for two standard benchmarks THUMOS14 and Charades. 
		Our method obtains excellent results compared to state-of-the-art methods that uses similar features and loss functions on THUMOS14 dataset. 
		Similarly, our  weakly supervised method is only 0.3\% mAP behind a state-of-the-art supervised method on challenging Charades dataset for action localization.
	\end{abstract}

	\section{Introduction}
\begin{figure*}[t]
 \centering
 \includegraphics[width=1.0\textwidth]{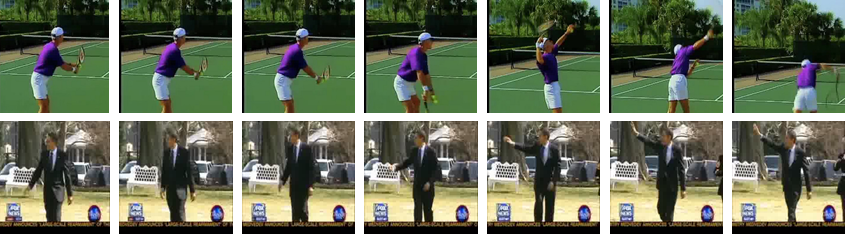}
 \caption{Example videos for action detection contain ``tennis-swing'' and ``hand-wave'' in top and bottom rows respectively. Labeling start and end frames of these actions requires not only a global understanding of these actions but also a local comparison of the candidate frames with their temporal neighbors in the same video. }
 \label{fig:Sample}
 \end{figure*}

	Action classification (\eg \cite{Bobick2001,Carreira2017,dollar2005behavior,Fernando2016,karpathy2014large,LaptevL03,Simonyan2014,wang2013action}) is an extensively studied problem in video understanding with important applications in surveillance, human-machine interaction and human behavior understanding. 
Recent advances in action classification can be attributed to powerful hierarchical learnable feature representations~\cite{karpathy2014large,Simonyan2014,Fernando2017}, introduction of large video datasets \cite{karpathy2014large, Kay2017}, the use of motion information (\eg optical flow~\cite{Simonyan2014}, dynamic images~\cite{Bilen2017}) and 3D convolutions~\cite{tran2014learning}. 
While recent methods such as \cite{Carreira2017,tran2018closer} have shown to obtain good action classification performance in various benchmarks, a remaining challenges in video understanding is to localize and classify human actions in long untrimmed videos. 
\begin{figure}[t]
\centering
\begin{tabular}{c}
\includegraphics[width=0.45\textwidth]{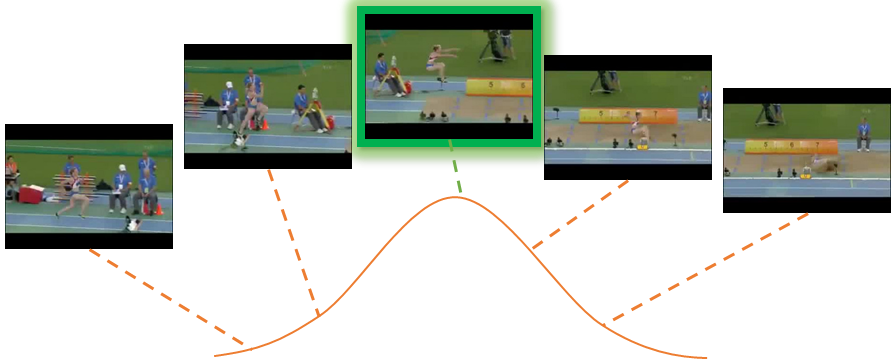}\\
(a) Video-specific (local) frame selection\\
\\
\includegraphics[width=0.45\textwidth]{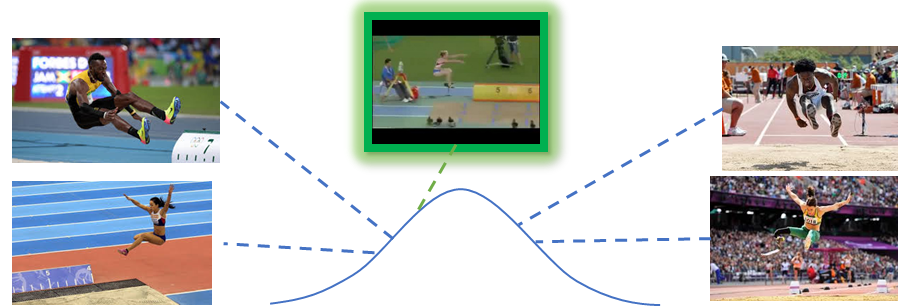}\\
(b) Dataset-specific (global) frame selection\\
\end{tabular}
\vspace{0.1cm}
\caption{Our method makes use of both video specific (\ie local) and dataset specific (\ie global) score prediction distributions to identify the most relevant set of frames for a given video in a weakly supervised manner for action localization and detection.  
The likelihood of a frame (shown in green) for an action category (``high-jump'') is obtained by comparing it to the other frames from the same video (local) and frames from other training videos (global) by using two Gaussian normalization functions.}
\label{fig:illustration}
\end{figure}
Recent methods \cite{Chao2018,zhao2017temporal} address localization of actions in long videos in a supervised manner and require action labels for each frame. 
The supervised paradigm has two shortcomings. First, frame labels are significantly more tedious and expensive to obtain than video-level labels. 
Second, temporal extent of actions are not as clear as spatial extents of objects (see~\Cref{fig:Sample} for an illustration). 
For instance, Sigurdsson et al.~\cite{sigurdsson2017actions} report only 58.7\% agreement for temporal boundaries of actions in MultiTHUMOS datasets among human annotators. 
To address these issues, we propose a novel \emph{weakly supervised} action detection method using only video-level labels. 
Our method is weakly supervised~\cite{zhou2017brief} because it does not use any frame-level labels during training, however, it outputs accurate frame-level labels at test time. 

Our model is trained to \textbf{select} and \textbf{classify} relevant frames for a given action using video-level labels.
It selects relevant frames using a deep neural network named \emph{frame selection module} which acts as a proxy for action localization. 
At the same time, it classifies relevant frames using another deep neural network, namely \emph{classification module}. 
Therefore, our model contains two modules (streams), one for frame selection \ie, frame selection module, and one for classifying each frame. 
Then our model fuses information from both modules to make a video level prediction and trains end-to-end using only video-level labels.

Inspired by the challenging task of predicting temporal boundaries of ``hand-wave'' action in the bottom row of~\cref{fig:Sample}, we hypothesis that accurate action prediction requires not only global understanding of an action class but also a closer look at the frames of a video and local comparison between its frames. 
We build the frame selection module on this idea such that it assigns a weight to each frame prediction based on both local and global statistics. 
The local one is realized by a local Gaussian function that picks the most likely frames for an action class and ignores outliers within a video. 
The predictions that have smaller and larger confidences with respect to the local mean prediction for an action class are normalized (regularized) using a local Gaussian function.
This operation is local because the selection involves analysis of predictions among the frames of a single video. 
This allows us to find the most robust set of predictions for a given action class within a video.

Similarly, the global frame selection strategy involves picking the most relevant frames for an action class from a video by comparing them to global action statistics that are learned over all the videos of that action class. 
This strategy selects frame predictions that are consistent with globally learned statistics.
The analysis of local and global modules are combined to obtain a joint distribution over frames and action classes (see~\cref{fig:illustration}). Finally the outputs of the frame selection and classification modules are combined.

In summary, our contributions are twofold: i) we propose a novel frame selection process for weakly supervised action localization using Gaussian normalization, 
ii) our Gaussian normalization of scores using both local and global statistics are effective for action localization. 
Our contributions result in a good improvement in action localization and detection in several challenging benchmarks obtaining results that are competitive with recent weakly supervised techniques.

	\section{Related Work}\label{s:related}
	\paragraph{Weakly supervised action localization} Weakly supervised action classification and localization has been studied in prior work~\cite{bojanowski2014weakly,bojanowski2015weakly,duchenne2009automatic,huang2016connectionist,Nguyen_2018_CVPR,Shou2018,Sigurdsson2017,Singh2017}. In~\cite{bojanowski2014weakly,bojanowski2015weakly,duchenne2009automatic}, the authors use movie scripts to obtain action labels and their approximate temporal boundaries from untrimmed videos and use them as a means of supervision to train action classifiers with various discriminative clustering algorithms. Duchenne~\etal~\cite{duchenne2009automatic} propose a discriminative learning formulation that simultaneously refines temporal action locations with classifier parameters. Bojanowski~\etal~\cite{bojanowski2014weakly,bojanowski2015weakly} extend \cite{duchenne2009automatic} by additionally exploiting the order of actions in a video clip to ensure that the classifier predictions are aligned with the orderings in the scripts. 

Extended Connectionist Temporal Classification~\cite{huang2016connectionist} utilizes weak annotations for action classification by aligning each frame with a label in a recurrent network framework. In contrast to \cite{huang2016connectionist} that learns from an ordered list of action labels per video, our method learns to localize action categories from weaker supervision, an unordered set of actions. In principle, such constraints can be incorporated to our learning formulation as constraints. 
A simple method that implicitly learns to find relevant parts of an object/action after randomly suppressing random parts of  images/videos is presented by Singh~\etal~\cite{Singh2017}. 
While this method is shown to be useful for preventing the network to focus only on discriminative segments, the final model does not achieve a better action classification performance.
A more effective weakly supervised action detection method that directly predicts the action boundaries using outer-inner-contrastive loss to parameterize classification loss in terms of temporal boundaries is presented by Shou~\etal~\cite{Shou2018}. 
Nguyen \etal~\cite{Nguyen_2018_CVPR} propose a loss function comprised of two terms that minimize the video-level action classification error and enforce the sparsity of the segment selection.

Recently Paul \etal~\cite{Paul2018} proposed to employ an attention-based mechanism to select relevant frames and apply pairwise video similarity constraints in a multiple instance framework. 
Liu \etal~\cite{Liu_2019_CVPR} also utilizes an attention module along with multiple classification streams, each can focus on different discriminative aspects of actions.
As a matter of fact, our model also consists of multiple specialized streams, however it differs significantly in terms of temporal modeling functions such as Gaussian and softmax normalization functions to select relevant frames in a weakly supervised manner. We compare to \cite{Singh2017,Shou2018,Nguyen_2018_CVPR,Paul2018,Liu_2019_CVPR} quantitatively in \cref{s:experiments}.

Wang~\etal~\cite{Wang2017} also employ a two stream method based on~\cite{Bilen2016a} for video action detection and localization.
Our method differs to Wang~\etal~\cite{Wang2017} as our method not only considers local video statistics but also global-dataset-specific score distributions which is crucial for accurate action localization. As also observed in~\cite{Nguyen_2018_CVPR}, the frame selection mechanism in \cite{Wang2017} is limited to select few examples due to the exponential term in softmax operator. While such a mechanism has been shown in \cite{Bilen2016a} to perform well when there are limited instances for object detection, it is not as effective to localize actions which typically comprised of many frames. 

\paragraph{Weakly supervised action segmentation}~\cite{Richard_2018_CVPR,Ding_2018_CVPR} is another closely related topic to weakly supervised action detection. It focuses on aligning dense textual descriptions (\eg recipes) such as action transcripts with the frames of the corresponding video (\eg cooking video) by predicting temporal boundaries between different actions. Richard~\etal~\cite{Richard_2018_CVPR} use context modeling with context free grammar to perform action segmentation. Ding~\etal~\cite{Ding_2018_CVPR} use a temporal convolutional feature pyramid to find coarse ground truth labels and a iterative refinement step using transcripts.

\paragraph{Weakly supervised object localization} learn to localize object instances spatially in still images from image-level labels only. The recent work in weakly supervised object detection propose better deep network architectures~\cite{Bilen2016a},  initialization~\cite{song2014learning}, learning strategies~\cite{cinbis2017weakly} that are less prone to over-fitting, use of various advanced cues such as abjectness~\cite{deselaers2010localizing}, object size~\cite{shi2016weakly} and co-occurrence~\cite{shi2017transfer}.
	\section{Problem, approach and model}\label{s:model}
	\begin{figure*}[t]
	\centering
	\includegraphics[width=0.8\textwidth]{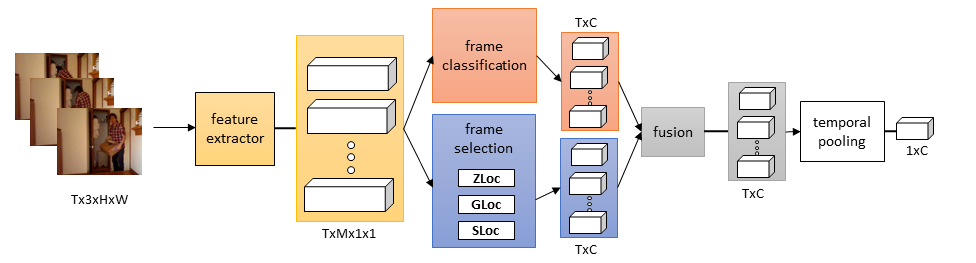}
	\caption{Illustration of our weakly supervised Gaussian action detection framework (WSGN).}
	\label{fig:archioverview}
\end{figure*}
In this section we present our Weakly Supervised Gaussian Network (\textbf{WSGN}) for action localization.
In~\cref{sec.method.prob}, we present our problem definition, and then in~\cref{sec.method.naive} we present a simple naive approach to weakly supervised action localization. Finally, in~\cref{sec.our.method}, we present our approach and the methodology.
\subsection{Weakly supervised action localization problem}
\label{sec.method.prob}
Let $V=\left< I_1, I_2, \cdots I_t, \cdots, I_{T} \right>$ be a sequence of frames where $I_t\in \mathcal{I}=\mathbb{R}^{3\times H\times W}$ is the $t^{th}$ frame of the video $V$. $T$ denotes the video length which vary from video to video. Assume that we are given a set of $N$ training videos and its video-level labels $\{V^i,\mathbf{y^i}\}$ where $\mathbf{y}\in \mathcal{Y} = \{0,1\}^C$ indicates the presence/absence of human action classes as a $C$-dimensional ground-truth binary vector for \emph{each training video} $V$. The $q^{th}$ element of the vector $\mathbf{y}$ is set to one when $q^{th}$ human action is present in the video, otherwise it is set to zero. We wish to learn a function that predicts the presence/absence of human action classes not at video-level but at frame-level for a testing video \ie to predict C-dimensional binary vector $\mathbf{y}_t$ for each frame $I_t$.
The learning becomes weakly supervised as what is being predicted at test time is more detailed than what is used for training~\cite{zhou2017brief}.
We predict frame label vector (\ie $\mathbf{y}_t$) at test time for each frame using a model that is trained with \emph{video-level labels} $\mathbf{y}$.
Therefore, our action localization task (\ie predicting $\mathbf{y}_t$ for each frame) is weakly supervised.

Let us denote a trainable feature extractor that returns a $M$-dimensional vector for each frame by $f(I_t,\theta):\mathcal{I}\rightarrow \Omega=\mathbb{R}^M$. 
Here $\theta$ are the learned parameters of $f$.
A classification network $h(\cdot,\theta_{cls}):\Omega\rightarrow\mathcal{Y}$ takes the feature vector $f(I_t,\theta)$ and returns a C-dimensional \emph{action classification score vector}. 
Here $\theta_{cls}$ are the trainable parameters of $h$. 
Action classification score vector for frame $I_t$ is then obtained by the joint model $h(f(I_t))$.
Next, we present a simple weakly supervised action localization method which we use as a baseline in our experiments.

\subsection{Naive weakly supervised action localization}
\label{sec.method.naive}
When frame-level action class annotation vectors $\mathbf{y}_t$ are known for the frames of training videos, one can train $f$ and $h$ to minimize binary cross-entropy loss at frame-level. As we assume that no ground truth frame-level labels are available for training, we are limited to use video-level label vectors $\mathbf{y}$ to train our action localization model that predicts $\mathbf{y}_t$ at test time. 

A simple strategy to obtain a video-level prediction $\hat{\mathbf{y}}$ from a sequence of frames is to average frame-level predictions over the whole sequence as follows:
\begin{equation}
\hat{\mathbf{y}} =  \sum_{t=1}^{T} \frac{1}{T} \sigma(h(f(I_t,\theta),\theta_{cls})
\label{eq.naive}
\end{equation} where $T$ is the number of frames in the sequence and can vary from video to video, $\sigma$ is the softmax normalization function over predicted score vector for each frame.
To train such a model, we minimize the binary cross entropy loss $\mathcal{L}(\mathbf{y}, \hat{\mathbf{y}})$ over predicted probability vector ($\hat{\mathbf{y}}$) and the ground truth.
During testing, we use function $h(f(I_t))$ to label each frame. However, this method naively considers an equal importance for each frame to obtain a video level score prediction by simply averaging their scores. We denote this approach the \textbf{Naive Weakly Supervised} action localization. Here we hypothesize that a good model should carefully choose the ``relevant'' frames for the action and then average the scores of only those. 

\subsection{Our approach}
\label{sec.our.method}
Our approach is to learn another network in addition to the ``classification module'' (\ie $\sigma(h(f(I_t,\theta),\theta_{cls})$) that can identify the relevance of each frame for the task of video action classification. 
We call this network ``frame selection module'' and denote it by function $g(I_t,V,\theta_g)$ where $\theta_g$ is the learnable parameter vector--see~\cref{fig:archioverview}. 
This network function acts as a proxy for action localization and weighs each frame per action class depending on the relevance of the frame to recognize the action. 

Similar to $h$, $g$ function also returns a C-dimensional weight vector for each frame $I_t$. However, 
$g$ function differs to the classification module function $\sigma \circ h \circ f$ in two aspects. 
First, its objective is to find frames that are relevant to each action class. Second, while the classification module $h$ scores each frame independent of other frames, the frame selection module scores each frame relatively by considering frames both from the video $V$ and the entire dataset. We describe the details of the relative scoring functions in the following paragraphs.

The final video-level prediction is obtained by a weighted average of all frame classification predictions where the weights are defined by $g$ as shown in equation~\ref{eq.ourbase}. Here $\odot$ is the element-wise product between weights and classification probability vectors. 
\begin{equation}
 \hat{\mathbf{y}} = \frac{1}{T} \sum_{t=1}^{T}  g(I_t,V,\theta_g) \odot \sigma(h(f(I_t,\theta),\theta_{cls}))
 \label{eq.ourbase}
\end{equation}
The video-level prediction $\hat{\mathbf{y}}$ can now be used with a binary cross entropy loss $\mathcal{L}(\mathbf{y}, \hat{\mathbf{y}})$ and enables our action localization method to be trained with video-level labels.
During inference, we simply skip this temporal averaging step and instead use $g(I_t,V,\theta_g) \odot \sigma(h(f(I_t,\theta),\theta_{cls})$ to obtain frame-level predictions and perform action localization by using these scores. In next part, we discuss how to formulate a good $g$ function for the task of action localization.

\paragraph{{WSGN}: Gaussian frame selection module.}
Here we explain the frame selection module $g$ which is complementary to the classification module $\sigma(h(f(I_t,\theta),\theta_{cls}))$.
To this end, we design the frame selection module in a way that it can predict the relevance of a frame by a comparative analysis to the rest of the frames.
In particular $g$ function consists of three components which are responsible for (i) extracting features from sequence of frames, (ii) predicting a C-dimensional score vector for each frame based on the extracted features, (iii) normalizing those score vectors to select frames.
For the first component, $g$ shares the feature extractor $f$ with the classification module for computational efficiency. For the second part, $g$ has a dedicated classifier also denoted by $h(\cdot,\theta_{det})$ which takes the feature $f(I_t)$ and returns a $C$-dimensional \emph{action selection score vector} but parameterized a different set of parameters $\theta_{det}$. 
As the input and output dimensions of $h(\cdot,\theta_{det})$ are same with $h(\cdot,\theta_{cls})$, we simply use the same function structure $h(\cdot,\cdot)$ for brevity.
The frame selection score vector obtained by $h(f(I_t,\theta),\theta_{det})$ is then denoted by $\mathbf{x}_t$ and the class-specific score for the $q^{th}$ action class is then denoted by scalar $\mathbf{x}^{q}_{t}$.

The objective of frame selection module $g(\cdot)$ is to select relevant frames for a given action.
To do so we make use of both video-specific and dataset-wide statistical information to find frame-action predictions that are most probable using \emph{normalization functions}.
These \emph{normalization functions} compare predicted scores for each frame against other frames to obtain a \emph{likelihood estimate} for each prediction $h(f(I_t),\theta_{det})$.
However, the aim here is not to obtain an estimate for the presence of action class $q$ in frame $I_t$ as done by the classification module $h(f(I_t),\theta_{cls})$ but to estimate a likelihood of each prediction $h(f(I_t),\theta_{det})$ with respect to local and global score distributions.
Higher the likelihood estimate of a prediction with respect to others, a higher weight is given by frame selection module, to the corresponding classification prediction in~\cref{eq.ourbase}.
Using the feature extractor $f(\cdot,\theta)$, frame selection function $h(\cdot,\theta_{det})$ and normalization functions, our frame selection module $g(\cdot)$ finds relevant frames for each action class. 
Next we describe three \emph{normalization functions} including local (ZLoc), global (GLoc) and softmax (SLoc) that are used with our WSGN method.

\paragraph{\textbf{ZLoc}: Local variant of WSGN} 
The local variant of WSGN model uses a Gaussian normalization function to find weights for each prediction using a likelihood estimate over the frames of a single video.
We estimate the likelihood of frame selection scores ${x}^{q}_{t}$ for $q^{th}$ class relative to all other frames within the same video $t = \{1,2,\cdots, T \}$ using a Gaussian likelihood function as follows:
\begin{align}
  \label{eq:znormpr}  
  z^{q}_{t} =  \exp({-[\frac{  {x}^{q}_{t} - \mu^{q}_{z}  }{ s^{q}_{z} }]^{2}}).
\end{align}
Here, $ \mu^{q}_{z}$ and $s^{q}_{z}$ are the statistical mean and the standard-deviation of all frame selection scores for $q^{th}$ class obtained by $h(f(\cdot,\theta),\theta_{det})$ for video $V$.
In fact $z^{q}_{t}$ is a probability estimate of ${x}^{q}_{t}$ with respect to all other frame selection scores within the video.
It assigns lower probabilities to those predictions that are very different from the mean prediction.
If ${x}^{q}_{t}$ is very large compared to the mean, it is considered as an outlying prediction. This encourages our model to identify not only the most salient frame for the action but also the whole temporal extent of the action.

By using \cref{eq:znormpr}, $g$ function assigns a class-specific weight ${z}^{q}_t$ to each frame $I_t$ which is further multiplied with the classification prediction (\ie from $\sigma(h(f(I_t,\theta),\theta_{cls}))$), as indicated in \cref{eq.ourbase}.
Let us denote this normalization operation by $g_{zloc}(\cdot)$ which takes all action selection score vectors ($\mathbf{x}_t$ for all $t = \{1,2,\cdots, T \}$) and returns a weight vector $\mathbf{z}_t$ for each frame.
Then the function $g(I_t,V,\theta_g)$ would return a weight vector $\mathbf{z}_t$ where
\begin{equation}
g(I_t,V,\theta_g) = g_{zloc}(\cdot) \circ h(\cdot,\theta_{det})  \circ f(\cdot,\theta)(I_t,V). 
\end{equation}
The $g_{zloc}(\cdot)$ function does not have any learnable parameters, thus learning the local variant of our method (denoted by ZLoc or $Z_{loc}$) involves optimizing  three sets of parameters $\theta$, $\theta_{cls}$ and $\theta_{det}$ using the sbinary cross-entropy loss.

\paragraph{\textbf{GLoc}: Global variant of WSGN.}
\label{sec.GLoc.method}
While ZLoc variant of our WSGN method considers statistics from frames of a single video to normalise scores, the global variant one \textbf{GLoc} compares each frame frame selection score ${x}^{q}_{t}$ with the frames from all the training videos.
As a direct comparison to all frames is computationally expensive and not even possible within the memory of the standard GPUs, we instead choose to use dataset-wide or global statistics with a Gaussian function per action category over the frame selection scores. To this end we propose to learn a mean vector ($\mu^{q}_{l}$) and standard deviation vector ($s^{q}_{l}$) per action category jointly along with the other network parameters. The subscript $l$ of $\mu^{q}_{l}$ is used to indicate that they are learned but not statistically computed over the scores.
Both $\mu^{q}_{l}$ and $s^{q}_{l}$ are learned using the training samples and therefore, representative of the global dataset specific information.
Our new GLoc normalization operation is then given by~\cref{eq:gnormpr}. 
\begin{equation}  
  l^{q}_{t} =  \exp({-[\frac{  {x}^{q}_{t} - \mu^{q}_{l}  }{ s^{q}_{l} }]^{2}})
  \label{eq:gnormpr}  
\end{equation}
This normalization function is denoted by $g_{gloc}(\cdot)$ which takes the class selection score vector $\mathbf{x}_t$ as input.
The weight vector returned by global GLoc approach is denoted by $\mathbf{l}_t = g(I_t,\theta_g)$ where 
\begin{equation}
g(I_t,\theta_g) = g_{gloc}(\cdot,\mathbf{\mu}_{l},\mathbf{s}_{l}) \circ h(\cdot,\theta_{det})  \circ f(\cdot,\theta)(I_t). 
\end{equation}
In contrast to local Gaussian approach (ZLoc), in GLoc, we learn parameter vectors $\mathbf{\mu}_{l}$ and $\mathbf{s}_{l}$ in addition to $\theta$, $\theta_{cls}$ and $\theta_{det}$.
The weight $\mathbf{l}_t$ is estimated not from a single video but all the training  samples. If the frame selection score ${x}^{q}_t$ is more likely w.r.t. global score distribution, then ${l}^{q}_t$ will be higher and the prediction from the classification path $\sigma(h(f(I_t,\theta),\theta_{cls}))$ for class $q$ is highly weighted.

\paragraph{\textbf{SLoc}: Softmax variant of WSGN.} 
For completeness, we also propose to use a commonly used normalization function, 
softmax but apply it to normalize the video specific scores $x^{q}_{t}$ over the frames of a video but not over the feature channels so that the sum of the frame selection scores are normalized to 1 for a video:
\begin{equation}  
s^{q}_{t} =  \frac{ e^{ x^{q}_{t} } } {\sum_{i=1} e^{ x^{q}_{i} }}
\label{eq:snormpr}  
\end{equation}
This normalization function is then denoted by $g_{sloc}$.

\paragraph{WSGN: Complete model.} We make use of all three normalization function, namely the local ZLoc, global GLoc and softmax-based SLoc variants in our WSGN.
To integrate the predictions from three streams, we propose a simple averaging strategy, i.e. $g = \text{avg}(g_{zloc}(\cdot), g_{gloc}(\cdot), g_{sloc}(\cdot)) \circ  h(\cdot,\theta_{det})  \circ f(\cdot,\theta)$,
%
where avg denotes element-wise averaging over three normalization functions.
The combined class-specific frame selection weight for frame $I_t$ can simply be obtained by the average of weights \ie $\frac{1}{3} (z^{q}_{t}+l^{q}_{t}+s^{q}_{t})$.
Now we can finally combine the predictions of classification module denoted by $\sigma(h(f(I_t,\theta),\theta_{cls}))$ and the frame selection module by $g(\cdot)$.
A visual illustration of our method is shown in~\cref{fig:archioverview}.
For action detection and localization, we use the score returned by $g(I_t,V,\theta_g) \odot \sigma(h(f(I_t,\theta),\theta_{cls})$ for each frame.

	\section{Experiments} \label{s:experiments}
	\subsection{Datasets}
\label{sec:datasets}
We evaluate our \textbf{WSGN} method on two standard action localization benchmarks, namely the Charades~\cite{Sigurdsson2016} and THUMOS14~\cite{THUMOS14}.

\paragraph{Charades}~\cite{Sigurdsson2016} is composed of 9,848 indoor videos with an average length of 30 seconds, comprising 157 action classes from 267 different people that are recorded in their homes and performing everyday activities. 
Each video is annotated with action labels and duration which allow evaluation for action localization. 
We use a standard evaluation procedure introduced in~\cite{Sigurdsson2017} for action localization using fixed train (7985) and validation (1863) splits. 
As done in~\cite{Sigurdsson2017}, we predict a score for 157 classes for 25 equally spaced time-points in each video and then report action localization mAP.

\paragraph{THUMOS14}~\cite{THUMOS14} dataset consists of very long videos (average length is 3.5 minutes) having 20 human action classes for action detection task. In this dataset, we follow the evaluation procedure in the previous work~\cite{Singh2017,Shou2018} to provide a fair comparison. Concretely, we use the validation set containing 200 untrimmed videos for training and evaluate our model on the test set containing 213 videos.

\subsection{Implementation details}

\emph{Features and networks:} We use two convolutional neural networks; namely the VGG16~\cite{Simonyan2014a} and ResNet34~\cite{He2016} that are pretrained for ImageNet classification task~\cite{russakovsky2015imagenet} for Charades dataset. These are trained end-to-end.
We use ImageNet pre-trained I3D \textbf{(I3D-I)} and UntrimmedNet~\cite{Wang2017} features for THUMOS14 experiments.
Furthermore, we evaluate our methods using I3D networks~\cite{Carreira2017} that is pretrained for video action classification on Kinetics dataset~\cite{Kay2017} (denoted by \textbf{(I3D-K)}) for both THUMOS14 and Charades experiments to obtain competitive state-of-the art results. These are only fine-tuned and not trained end-to-end.
For all these networks, we take the output before the classification layer and include a dropout layer with a dropout rate of 0.5 for image classification networks and a dropout rate of 0.8 for I3D networks. 
These serve as our feature extractor network $f(,\theta)$.
As the feature classification networks (\ie $h(,\theta_{cls})$ and $h(,\theta_{det})$), we use a simple two layered neural network to produce classification scores (hidden layer size is set to the input feature size).

\emph{Training details:} We use a learning rate of $10^{-4}$ for ResNet34, and I3D and a learning rate of $10^{-3}$ for VGG and trained for a maximum of 80 epochs. 
We use a batch size of 128 videos and 32 sub-batches and a weight decay of 0.0005.
Because some videos are very long, we sample every $5^{th}$ frame and perform a temporal data augmentation (vary the start of the sampled sequence from 1st to 15th frame).
Only for THUMOS14, during inference, we use all frames, however we set the mini-batch size to one to make sure we fit videos in GPU memory of (4$\times$16GB).
We use standard, data augmentation at video frame level (flipping, random cropping, etc.) but apply the same augmenting operation for the entire video to obtain a temporally smooth video after data augmentation.

%
\subsection{Ablation study on Charades dataset.}

In this section we compare several baselines and variants of weakly supervised models presented in~\cref{s:model} on Charades dataset. 
We experiment with three network architectures, namely VGG16, Resnet34 and I3D-K.
For I3D-K, we use bi-linear interpolation to obtain frame-wise feature representation and fine-tune with video level annotations.

We analyze the impact of different normalization functions \ie ZLoc, GLoc and SLoc of our WSGN.
We compare our method with (1) RGB-based naive weakly supervised baseline (\textbf{Naive}) which corresponds to~\cref{eq.naive}.
(2) We also report results for weakly supervised action localization only with Softmax normalization denoted by \textbf{SLoc}, (3) weakly supervised action localization only with local Gaussian normalization denoted by \textbf{ZLoc} (4) weakly supervised action localization only with global parametric Gaussian normalization denoted by \textbf{GLoc}. 
Our complete model shown in~\Cref{fig:archioverview} is denoted by \textbf{WSGN-Complete}.  
We also report results for fully \textbf{Supervised} case where we train a model using frame level annotations. 
In this case, we minimize the a combinations of losses $\mathcal{L}(\mathbf{y},\hat{\mathbf{y}} ) + \sum_t \frac{1}{T} \mathcal{L}(\mathbf{y}_t, \hat{\mathbf{y}_t} )$ where $\hat{\mathbf{y}} = \frac{1}{T} \sum_t \sigma(h(f(I_t,\theta),\theta_{cls})$ and $\hat{\mathbf{y}_t} = \sigma(h(f(I_t,\theta),\theta_{cls})$.
Results are shown in~\Cref{tab:tab-ablation}.
\begin{table}
\begin{center}
\scriptsize{
\begin{tabular}{lccc}
\toprule
& VGG16 & Resnet34 & I3D-K \\
\midrule
Supervised          &   9.0 &  10.1 &   18.7 \\
\midrule
\nwss               &   5.2 &   5.2 &   13.7 \\
WSGN - SLoc 		&   6.0 &   7.1 &   14.9 \\
WSGN - ZLoc 		&   8.7 &   9.0 &   16.8 \\
WSGN - GLoc 		&   8.5 &   9.3 &   17.2 \\
WSGN - SLoc + GLoc 	&   8.7 &   9.4 &   18.2 \\
WSGN - ZLoc + GLoc 	&   8.8 &   9.4 &   18.2 \\
WSGN - Complete &   \textbf{8.9} & \textbf{9.7}	&   \textbf{18.3}  \\ 
\bottomrule
\end{tabular}
}
\end{center}
\caption[Caption for ablation]{Action localization performances in mAP(\%) for supervised and weakly supervised methods with different base networks on Charades by using only RGB input. WSGN - SLoc: Softmax normalization, WSGN - ZLoc: local Gaussian normalization, WSGN- GLoc: global parametric Gaussian normalization. \footnote{Note that I3D supervised results are lower than what is reported in the Charades competition~\url{http://vuchallenge.org/charades.html}. This is because the competition setup is different where they use all train, validation data to train and test on 2000 test videos.}}
\label{tab:tab-ablation}
\end{table}

Several interesting observations can be listed based on these results. 
First, we obtain considerable improvements in action localization using our WSGN-Complete method over naive approach (improvement of \textbf{3.7} mAP for VGG16, \textbf{4.5} mAP for Resnet34, and \textbf{4.6} mAP for I3D).
Interestingly, our method seems to gain more when much richer network architectures are used as the best improvement over \nws~is obtained with I3D network.
Secondly, as an individual method, global Gaussian-based WSGN-GLoc seems the most effective one.
Local Gaussian-based normalization method (ZLoc) is also as effective as GLoc method.
Because SLoc uses softmax selection, its output is sparse and thus it could select only few important frames rather than entire coverage of an action. This leads to relatively low performance for Sloc.
Combination of both ZLoc and GLoc is considerably more effective.
Combination of all normalization methods seems to be the most beneficial in this dataset for action localization.
Our weakly supervised results are surprisingly as good as the fully supervised method on all three network architectures indicating the effectiveness of our novel frame selection module.
We conclude that WSGN weakly supervised action localization is effective on Charades dataset across wide variety of base network architectures.
\subsection{Ablation study on THUMOS14.}
\label{sec.thumos}
In this section we evaluate our method on THUMOS14 dataset using ImageNet pretrained I3D (I3D-I).
We extract features from both RGB and optical flow streams following the protocol used in~\cite{Carreira2017} and then use bi-linear interpolation to obtain frame-wise feature representation and fine-tune with video level annotations.
We evaluate varying the detection IoU threshold using the standard evaluation protocol as in~\cite{THUMOS14}. 
As our method is only weakly supervised, to generate action detection boundaries \ie start and end of each action, we make use of two heuristics. 
First, we threshold scores to find candidate frames and then generate candidate segments using those consecutive frames that has a score greater than the threshold. 
It should be noted, because we use local/global Gaussian normalization, returned scores are already managed to get rid of outliers and very small scores.
Secondly, we use only those candidate segments that are longer than one second for evaluation. 
We report results in~\Cref{tab:tab-ablation.det}.

\begin{table}
\begin{center}
\scriptsize{
\begin{tabular}{lccccc}
\toprule
Method & 0.1 & 0.2 & 0.3 & 0.4 & 0.5  \\
\toprule
Supervised & 59.7 & 51.9 & 47.4 & 40.1 & 32.8 \\ \hline
\nwss&  42.9&  36.6&  28.0&  20.9&  14.2 \\
WSGN - SLoc&  45.3&  38.9&  31.0&  23.5&  16.3 \\
WSGN - ZLoc&  54.5&  47.8&  38.8&  28.9&  20.0 \\
WSGN - GLoc&  45.2&  38.8&  30.2&  22.4&  14.6 \\
WSGN - SLoc + GLoc&  48.4&  42.5&  34.1&  26.0 &  18.0 \\
WSGN - ZLoc + GLoc&  54.7&  \textbf{48.6}&  \textbf{39.4}&  29.4&  20.7 \\
WSGN - Complete&  \textbf{55.3}&  47.6&  38.9&  \textbf{30.0}&  \textbf{21.1} \\ \hline
Gap & 4.4 & 4.3 & 8.5 & 10.1 & \textbf{11.7} \\
Improvement & \textbf{12.4} & 11 & 10.9 & 9.1 & 6.9 \\
\bottomrule
\end{tabular}
}
\end{center}
\caption{Action detection performance on THUMOS14 dataset using I3D-I for variants of our weakly supervised WSGN action detection method. We change IoU threshold from 0.1 to 0.5 and report results. WSGN - SLoc: Softmax normalization, WSGN - ZLoc, local Gaussian normalization, WSGN- GLoc: global parametric Gaussian normalization. The Gap between the supervised results and our WSGN (SLoc + ZLoc + GLoc) is shown. The improvement obtained over \nws~method is also shown. }
\label{tab:tab-ablation.det}
\end{table}
We observe a similar trend to Charades dataset where all variants of our WSGN is effective than \nws~method.
Interestingly, even if our supervised results are far better than our weakly supervised method, obtained results are very encouraging.
The improvement we obtain over \nws~method is comparatively greater than the gap between supervised performance and our WSGN (SLoc + ZLoc + GLoc).
We conclude that our method is very effective for weakly supervised action detection in THUMOS14 dataset.


%
%
%
\subsection{Comparison to prior state-of-the art.}
We compare with several weakly supervised action detection methods that have been evaluated on THUMOS14 dataset~\cite{Sun2015,Singh2017,Wang2017,Shou2018,Nguyen_2018_CVPR,Paul2018,Liu_2019_CVPR} in~\Cref{tab:tab.soa.det}.
In particular, some very successful recent methods such as STPN~\cite{Nguyen_2018_CVPR}, W-TALC~\cite{Paul2018} and Comp~\cite{Liu_2019_CVPR} use I3D pretrained on Kinetics (I3D-K).
Therefore, we report results with I3D-K and UntimmedNet~\cite{Wang2017} features (UNTF).
Additionally, we make use of CASL loss presented in~\cite{Paul2018} to further improve our results. 
\begin{table}
\begin{center}
\resizebox{\columnwidth}{!}{%
\begin{tabular}{llccccccc}
\toprule
& Method & 0.1 & 0.2 & 0.3 & 0.4 & 0.5 & 0.6 & 0.7 \\
\toprule
&Supervised~faster-rcnn~\cite{Chao2018} & 59.8 & 57.1 & 53.2 & 48.5 &  42.8 & 33.8 & 20.8 \\ \midrule
\multirow{12}{*}{\rotatebox{90}{Weak Super.}} 
& LAF~\cite{Sun2015} & 12.4 & 11.0 & 8.5 & 5.2 & 4.4 & -- & -- \\
&Hide-and-seek~\cite{Singh2017} & 36.4 & 27.8 & 19.5 & 12.7 & 6.8 & -- & -- \\
&UntrimmedNets~\cite{Wang2017} & 44.4 & 37.7 & 28.2 & 21.1 & 13.7 & -- & -- \\
&AutoLoc~\cite{Shou2018} & -- & -- & 35.8 & 29.0 & 21.2 & 13.4 & 5.8 \\ 
&STPN~\cite{Nguyen_2018_CVPR} (I3D-K) & 52.0 & 44.7 & 35.5 & 25.8 & 16.9 & 9.9 & 4.3 \\
&W-TALC~\cite{Paul2018}  (I3D-K) &  55.2 & 49.6 & 40.1 & 31.1 & 22.8 & & 7.6 \\ 
&Comp-~\cite{Liu_2019_CVPR} (I3D-K) & 57.4&50.8&41.2&32.1&23.1&15&7.0\\
&Our (I3D-K) & 55.3 & 48.8 &37.2 &30.2 &21.1 &13.8 &8.2 \\
&Our+CASL (I3D-K) &\textbf{57.9}&\textbf{51.2}& \textbf{42.0}& \textbf{33.1}&\textbf{25.1}&\textbf{16.7}&\textbf{8.9} \\
\cline{2-9}
&W-TALC~\cite{Paul2018} (UNTF) &  49.0 & 42.8 & 32.0 & 26.0 & \textbf{18.8} & -- & 6.2 \\
&WSGN (UNTF) & \textbf{51.1}&  \textbf{44.4} &  \textbf{34.9}&  \textbf{26.3}&  18.1 & \textbf{11.6} & \textbf{6.5} \\

\bottomrule
\end{tabular}
}
\end{center}
\caption{Action detection performance on THUMOS14 dataset for various weakly supervised state-of-the-art methods.
We also show results for supervised state-of-the-art method~\cite{Chao2018} as a reference.
We also compare with s.o.a. W-TALC~\cite{Paul2018} with UntimmedNet~\cite{Wang2017} features (UNTF).
}
\label{tab:tab.soa.det}
\end{table}
Our method trained with the CASL loss outperforms all other methods.
Also our method without CASL loss is better than recent effective methods such as STPN~\cite{Nguyen_2018_CVPR} which relies on feature attention over frame features similar to us.
Indeed, the idea of co-activity loss presented in W-TALC~\cite{Paul2018} is effective and complimentary to our method.
In-fact, state-of-the art methods such as recently presented~\cite{Liu_2019_CVPR} might be complimentary to us as well.
We leave the use of diversity loss~\cite{Liu_2019_CVPR} for future work.
We obtain better results with UNTF features compared to~\cite{Paul2018}. 
Especially, for IoU of 0.1, our method is only \textbf{1.9 mAP} behind state-of-the-art faster-rcnn-temporal-localization~\cite{Chao2018}. 
However, supervised methods perform way better than weakly supervised methods for larger IoU thresholds.
This is not surprising as the task becomes difficult with larger IoUs.
Nevertheless, we obtain somewhat encouraging results even compared to state-of-the-art supervised methods for smaller IoU thresholds while obtaining some encouraging results even for very large IoU of 0.7--see~\Cref{tab:tab.soa.det}.

To the best of our knowledge, no prior weakly supervised method have evaluated on challenging Charades dataset for action localization. 
Therefore, we only compare with supervised methods, which actually used frame level annotations. 
Results are reported in~\Cref{tab-soa}.
\begin{table}[t]
\centering
\scriptsize{
\begin{tabular}{llllr}
\toprule
& Method & Network & Input &  Loc.\\
\toprule
\multirow{6}{*}{\rotatebox{90}{Supervised}} &Temporal Fields~\cite{Sigurdsson2017}   & VGG16 & RGB		& 9.0\\
& Two Stream++~\cite{Simonyan2014}        & VGG16 & RGB+OF  	& 10.9\\ 
&Temporal Fields~\cite{Sigurdsson2017}   & VGG16 & RGB+OF 	& 12.8\\ 
&Super-Events~\cite{Piergiovanni2018}    & I3D   & RGB	  	& 18.6 \\ 
&Super-Events~\cite{Piergiovanni2018}    & I3D 	 & RGB+OF	& 19.4 \\ \midrule
\multirow{2}{*}{\rotatebox{90}{Wk. Sup.}} &WSGN (ours)     & VGG16 & RGB & 8.9 \\
&WSGN (ours)                  					& ResNet34 & RGB  & 9.7\\
&WSGN (ours)                  					&I3D & RGB &  18.3\\
\bottomrule
\end{tabular}
}
\vspace{0.1cm}
\caption{Comparison to the state-of-the action localization methods on Charades dataset.
All other methods  may use frame annotations of Charades dataset during the training process, hence fully supervised. 
Our method is only ``Weakly Supervised'' (WS).}
\label{tab-soa}
\end{table}
\begin{figure}
	\centering
	\includegraphics[width=0.99\columnwidth]{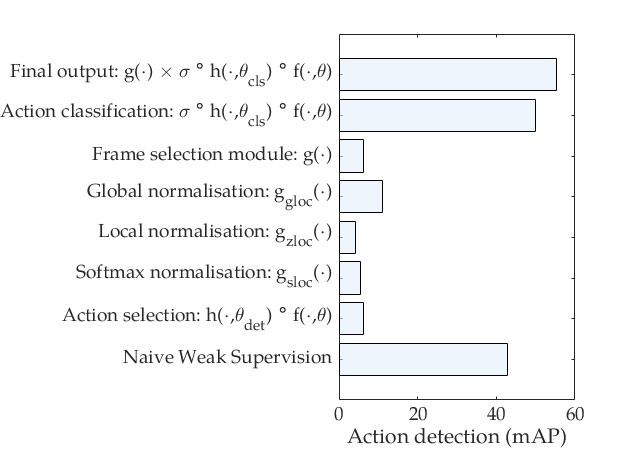}
	\caption{Dissecting the model outputs for action detection.}
	\label{fig:analysis}
\end{figure}
Notably, our method obtains competitive results compared to fully supervised methods using both VGG16 and I3D architectures.
Effective Temporal Fields~\cite{Sigurdsson2017} method obtains 9.0 mAP using VGG16 and RGB stream while our weakly supervised method is slightly worse (8.9 mAP).
Similarly, our WSGN with I3D feature extractor performs only \textbf{0.3 mAP} worse than supervised Super-Events~\cite{Piergiovanni2018} method.
This is an indication of the impact of our novel Gaussian normalization-based frame selection module.
However, our results are 1.1 mAP lower than Super-Events~\cite{Piergiovanni2018} when used with both RGB and optical flow (although we don't use optical flow for Charades dataset).
We conclude that our weakly supervised method is effective for video action localization (Charades dataset) and detection (THUMOS14 dataset).
\subsection{Analysis of localization components.}
\label{sec:analysis}
Our \name~method has several computational outputs that predicts categorical information.
For example the action classification score $h(,\theta_{cls}) \circ f(,\theta)$, and action selection score $h(,\theta_{det}) \circ f(,\theta)$. 
The global normalization function outputs $g_{gloc}()$, the local normalization outputs $g_{zloc}()$ and softmax normalization  outputs $g_{sloc}()$ also returns C-dimensional weights. 
Overall, the frame selection module $g(\cdot)$ outputs C-dimensional weight vector for each frame by taking average of all normalization functions.
Finally, both classification and frame selection module $g(\cdot)$ outputs are multiplied to get the frame-wise final prediction which we use for action localization. 
We perform an analysis on these outputs and report action detection performance in~\Cref{fig:analysis} using THUMOS14 dataset.

First, we see that all normalization outputs do not perform as good as classification output $h(,\theta_{cls}) \circ f(,\theta)$.
This is not surprising as the goal of normalization is simply frame selection for each action.
However, the classification module outputs performs satisfactorily and already obtains better results than \nws~method indicating in-fact the addition of frame selection module itself helped to improve the classification parameters as well.
This indirectly indicates that, our frame selection module $g(\cdot)$ help to improve the video representation $f(\cdot,\theta)$.
The best results are obtained by the final output indicating the advantage of having two separate functions for frame selection and classification.

	\section{Conclusion}\label{s:conclusion}
	In this paper we propose a weakly supervised action localization method where both frame selection and classification are learned jointly with a deep neural network in an end-to-end manner. We show that accurate action localization require both video-level and dataset-wide frame comparison. As demonstrated in our experiments, combination of both local and global strategies result in better performance and obtains state-of-the-art results in two challenging datasets. Importantly our method further narrows down the gap between the supervised and weakly supervised paradigms. For future work we plan to extend our weakly supervised localization method from temporal to spatio-temporal domain by exploring higher dimensional normalization strategies.

	\\
	\noindent
\textbf{Acknowledgment}: This research was supported by the National Research Foundation, Singapore (Award Number: NRF2015-NRF-ISF001-2451), the National Research Foundation Singapore under its AI Singapore Programme (Award Number: AISG-RP-2019-010), and the Agency for Science, Technology and Research (A*STAR) under its AME Programmatic Funding Scheme (Project \#A18A2b0046).
	
	{\small
		\bibliographystyle{plain}
		\bibliography{wacv}
	}
	
\end{document}